# Beyond Lipschitz: Data-Driven Robustness via Discrete Modulus of Continuity

**Jürgen Dölz**
Institute for Numerical Simulation
University of Bonn
Germany
doelz@ins.uni-bonn.de

**Michael Multerer** **Michele Palma**[*]
Dalle Molle Institute for Artificial Intelligence
USI Lugano
Switzerland
{michael.multerer, michele.palma}@usi.ch

## Abstract

Robustness of neural networks is commonly quantified via local or global Lipschitz constants. However, Lipschitz continuity can be overly coarse or overly restrictive as a global robustness measure, failing to capture nuanced, data-dependent behavior. We propose a data-driven, architecture-agnostic framework based on the discrete modulus of continuity (DMOC), a nonlinear generalization of Lipschitz continuity that provides a finer notion of robustness. Unlike many existing approaches, DMOC does not require access to model internals and instead evaluates regularity relative to the data distribution. This shifts the focus from the model to the data, which provide a data-driven baseline of regularity against which the network's robustness is assessed. We establish convergence results for DMOC-induced seminorms with explicit data-driven rates in terms of the separation distance, and introduce a scalable minibatch algorithm that reduces the quadratic cost of exact computation, enabling application to large-scale data sets such as ImageNet. Empirically, DMOC serves as an architecture-independent diagnostic: it distinguishes trained from untrained networks, reveals underfitting and overfitting regimes, and yields, as a special case, tight Lipschitz estimates comparable to state-of-the-art methods such as ECLipsE and ECLipsE-Fast.

## 1 Introduction

Deep neural networks are known to be sensitive to input perturbations, as highlighted by adversarial examples [Szegedy et al., 2014, Goodfellow et al., 2015]. In this context, the Lipschitz constant has emerged as a principled tool for certifying the robustness of a neural network. By bounding the maximum change of a model's output in response to a certain change of input, the Lipschitz constant provides theoretical guarantees on stability and enables the derivation of robustness certificates against adversarial perturbations [Weng et al., 2018]. Several works have focused on the certified approximation of the Lipschitz constant of neural networks [Virmaux and Scaman, 2018, Weng et al., 2018, Fazlyab et al., 2019, Latorre et al., 2020, Xu and Sivaranjani, 2024] or empirical estimation [Weng et al., 2018, Nazarov et al., 2025] and regularization methods enforcing a small Lipschitz constant after training [Gouk et al., 2021, Miyato et al., 2018] have been developed.

Unfortunately, using the Lipschitz constant for certifying the robustness of a neural network comes with two major drawbacks. First, many existing approaches rely on specific network architectures or come with limited theoretical guarantees. Second, while enforcing Lipschitz continuity in neural networks can address the issue of adversarial examples, it can also have a negative impact on the expressiveness and accuracy of neural networks. For example, consider the square root function $x \mapsto \sqrt{x}$ on $[0, 1]$, which is known to be continuous, but not Lipschitz continuous. Learning this function under a Lipschitz constraint may significantly reduce approximation efficacy. Even for Lipschitz continuous target functions, the accuracy of Lipschitz constrained neural network approximations depends strongly on the chosen constraints, and conservative constraints may degrade accuracy. On the other hand, the function $x \mapsto \tanh(x)$ on $[-100, 100]$ has Lipschitz constant $1$, but Lipschitz continuity overestimates variation on a global scale. Thus, learning it from data may benefit from additional regularization beyond Lipschitz constraints. These and similar observations and benchmarks on more involved examples have led to intense discussions in the community Tsipras et al. [2019], Huster et al. [2018], Béthune et al. [2022]. Thus, the Lipschitz constant as a measure of robustness can be either too coarse or too restrictive. A black-box, multi-scale notion of robustness with error guarantees that applies across architectures is still missing.

[*] Authors are listed in alphabetical order.

We propose the discrete modulus of continuity (DMOC) as a data-driven robustness measure for neural networks. The DMOC is a discrete approximation to the well-known modulus of continuity (MOC), which can be considered as a non-linear and more informative refinement of Hölder continuity. In particular, the Lipschitz constant of a Lipschitz continuous function can be recovered directly from its MOC. Moreover, every continuous function defined on a compact domain admits a MOC. Thus, neural networks, which are often designed to be continuous, admit one as well. Relying on the DMOC shifts the focus from model regularity to data regularity, providing a data-intrinsic baseline against which we assess the network's robustness. Our key contributions are:

**A data-centric robustness framework:** The DMOC provides a multi-scale, data-intrinsic notion of regularity that depends only on input-output behavior and applies to arbitrary architectures in a black-box setting.

**Theory:** We establish convergence results for DMOC-induced seminorms with explicit rates in terms of separation distance and fill distance, connecting the discrete estimator to its continuous counterpart.

**Scalability:** We propose a minibatch approximation that mitigates the quadratic cost of exact computation and enables evaluation on large data sets such as ImageNet.

**Empirical analysis:** We demonstrate empirically that DMOC offers a richer multi-scale description of network regularity than the Lipschitz constant and we propose alignment metrics for automated comparison of DMOCs. Moreover, our framework recovers, as a special case, tight Lipschitz estimates competitive with ECLipsE and ECLipsE-Fast.

This work is organized as follows. We first introduce the MOC and the DMOC and establish their connection to classical Lipschitz continuity via associated seminorms. We then prove an abstract convergence result for these seminorms for any non-decreasing weight function, with explicit rates in terms of the separation distance and fill distance. Building on this analysis, we present both an exact algorithm for computing DMOC and a scalable minibatch approximation for efficient evaluation on large data sets. We present numerical experiments that analyze characteristic DMOC patterns of trained and untrained networks and demonstrate how DMOC distinguishes underfitting from overfitting.

## 2 Methodology

### 2.1 Modulus of continuity

Let $\mathbb{R}^{n_1}$ and $\mathbb{R}^{n_2}$ be equipped with the geodesic metrics $d_X$ and $d_Y$, respectively. Further, let $D \subset \mathbb{R}^{n_1}$ be a convex and compact Lipschitz region with diameter $T_D := \operatorname{diam}(D) < \infty$. On $D$, we consider a continuous mapping $f\colon D \to \mathbb{R}^{n_2}$. The *modulus of continuity* (MOC) $\omega(f, \cdot)\colon [0,\infty) \to [0,\infty)$ of $f$ is given by

$$\omega(f,t) := \sup_{\substack{\boldsymbol{x},\boldsymbol{x}' \in D:\\ d_X(\boldsymbol{x},\boldsymbol{x}') \le t}} d_Y\big(f(\boldsymbol{x}), f(\boldsymbol{x}')\big).$$

In the setting under consideration, it is well known Lebesgue [1909], DeVore and Lorentz [1993] that the MOC has the following properties.

- The mapping $t \mapsto \omega(f,t)$ is continuous for all $t \ge 0$ with $\omega(f,0) = 0$.
- The mapping $t \mapsto \omega(f,t)$ is monotonically increasing in $t \ge 0$.

- The MOC is subadditive, i.e., $\omega(f, t_1 + t_2) \leq \omega(f, t_1) + \omega(f, t_2)$ for any $t_1, t_2 \geq 0$.

**Remark 2.1.** *The MOC allows to measure Lipschitz, Hölder, and even weaker regularity. An example of a non-Lipschitz and non-Hölder continuous function is $f\colon [0,1] \to [0,1]$ with $f(0) = 0$ and $f(x) = |\log x - 2|^{-1}$ for $x \in (0,1]$. Its MOC is $\omega(f,t) = f(t)$.*

Letting $L := \sup_{t>0} t^{-1}\omega(f,t) \in [0,\infty]$, the function $t \mapsto Lt$ is the least homogeneous linear majorant of $\omega(f,t)$. In particular, if $L < \infty$, then $f$ is Lipschitz continuous with the Lipschitz constant $L$. More generally, the MOC $\omega(f,\cdot)$ provides a quantitative, scale-dependent description of the sensitivity of $f$ to perturbations in its argument and thus serves as a general robustness measure encompassing Hölder- and Lipschitz continuity as special cases. As a rule of thumb, the faster the decay of the MOC towards zero, the more regular the function [Dölz and Multerer, 2025].

**Remark 2.2.** *In the particular case of feedforward neural networks, one considers mappings of the type $f = f_L \circ f_{L-1} \circ \cdots \circ f_1$ where each $f_\ell$ is Lipschitz continuous with Lipschitz constant $L_\ell$. A standard choice is $f_\ell(\boldsymbol{x}) = \phi_\ell(\boldsymbol{W}_\ell \boldsymbol{x} + \boldsymbol{b}_\ell)$, where $\boldsymbol{W}_\ell, \boldsymbol{b}_\ell$, $\phi_\ell$ denote weight matrix, bias and a Lipschitz continuous activation function, respectively. Then $f$ is Lipschitz continuous and an upper, but often loose, bound to its Lipschitz constant given by $L \leq \prod_{\ell=1}^{L} L_\ell$, see also Szegedy et al. [2014].*

## 2.2 Discrete modulus of continuity

We assume that $f$ is only known from samples $\{(\boldsymbol{x}_1, \boldsymbol{y}_1), \ldots, (\boldsymbol{x}_N, \boldsymbol{y}_N)\} \subset D \times \mathbb{R}^{n_2}$, giving rise to the *site-to-value map*

$$f_N\colon X_N \to Y_N, \quad \boldsymbol{x}_i \mapsto \boldsymbol{y}_i.$$

Herein, $X_N := \{\boldsymbol{x}_1, \ldots, \boldsymbol{x}_N\}$ denotes the set of *data sites* and $Y_N = \{\boldsymbol{y}_1, \ldots, \boldsymbol{y}_N\}$ the set of *data values*. As usual, we denote the *separation distance* and the *fill distance* of $X_N$ by

$$q_N := \min_{\boldsymbol{x}_i, \boldsymbol{x}_j \in X_N : \boldsymbol{x}_i \neq \boldsymbol{x}_j} d_X(\boldsymbol{x}_i, \boldsymbol{x}_j) \quad \text{and} \quad h_{D,N} := \sup_{x \in D} \min_{\boldsymbol{x}_i \in X_N} d_X(\boldsymbol{x}, \boldsymbol{x}_i),$$

respectively.

In analogy to the MOC, we introduce the *discrete modulus of continuity* (DMOC) as

$$\omega_N(f,t) = \omega_N(f_N,t) := \sup_{\substack{\boldsymbol{x}, \boldsymbol{x}' \in X_N \\ d_X(\boldsymbol{x}, \boldsymbol{x}') \leq t}} d_Y\big(f_N(\boldsymbol{x}), f_N(\boldsymbol{x}')\big),$$

As for to the continuous case, the slope of the least homogeneous linear majorant of $t \mapsto \omega(f_N, t)$ coincides with the Lipschitz constant of $f_N$, which we will thus consider as an approximation to the Lipschitz constant of $f$.

Due to the subadditivity, we immediately obtain a bound on the approximation of the DMOC in terms of the growth of the modulus of continuity close to zero.

**Theorem 2.3.** *Let $t > 0$ be fixed and let $X_N$ be an $r/2$-net for $0 < r \leq t$, i.e., $\bigcup_{n=1}^{N} B_{r/2}(\boldsymbol{x}_n) = D$. Then, there holds $0 \leq \omega(f,t) - \omega_N(f,t) \leq 3\omega(f,r)$.*

*Proof.* The proof follows immediately from [Dölz and Multerer, 2025, Theorem 3.6] by employing the subadditivity of $\omega(f,t)$, which yields $\omega(f,t) - \omega(f,r) \leq \omega(f,t-r)$. □

For a monotonically increasing function $\rho\colon [0,\infty] \to [0,\infty]$, we introduce the *$\rho$-normalized modulus of continuity* $\nu_\rho(f,t) := \rho(t)^{-1}\,\omega(f,t),\ t > 0$, and define the usual seminorm

$$|f|_\rho := \sup_{t>0} \nu_\rho(f,t). \tag{1}$$

For $\nu_{\rho,N}(f,t) := \rho(t)^{-1}\,\omega_N(f,t),\ t > 0$, the discrete analogue of the seminorm is given by

$$|f_N|_{\rho,N} := \max_{t \in Z_N \setminus \{0\}} \nu_{\rho,N}(f_N, t)$$

where $Z_N := \big\{d_X(\boldsymbol{x}_i, \boldsymbol{x}_j) : \boldsymbol{x}_i, \boldsymbol{x}_j \in X_N\big\}$. The discrete seminorm approximates the seminorm from below, i.e., for $X_N \subset X_{N'}$, $N' \geq N$, there holds $|f_N|_{\rho,N} \leq |f_{N'}|_{\rho,N'} \leq |f|_\rho$, see [Dölz and Multerer, 2025, Lemma 2.8]. In the following theorem, we quantify the speed of convergence.

**Theorem 2.4.** *Let $X_N$ be an $r/2$-net for $D$ with $r < T_D$. Let $\rho \geq 0$ be nondecreasing. Then, there holds*

$$0 \leq |f|_\rho - |f_N|_{\rho,N} \leq \omega\big(\nu_\rho(f,\cdot), 2r\big) + 4\rho(q_N)^{-1}\omega(f,r).$$

The proof can be found in Appendix A. As an illustration of Theorem 2.4, consider $f\colon [0,1] \to [0,1]$, $f(t) = t^\alpha$, $\alpha \in (0,1]$, where $[0,1]$ is equipped with the Euclidean metric. Choose $\rho(t) = t^\beta$, $\beta \in (0,\alpha)$. Then, there holds $\omega(f,t) = t^\alpha$, $\nu_\rho(f,t) = t^{\alpha-\beta}$ and $\omega(\nu_\rho(f,\cdot),t) = t^{\alpha-\beta}$. On a quasi-uniform $r/2$-net $X_N$ for $[0,1]$, we have $q_N \sim h_{D,N}$, and the theorem implies

$$0 \leq |f|_\rho - |f_N|_{\rho,N} \leq Cr^{\alpha-\beta}$$

for some constant $C > 0$ independent on $r$, as one would expect.

### 2.3 Computation of the DMOC

Given geodesic metrics $d_X, d_Y$ on $\mathbb{R}^{d_1}$ and $\mathbb{R}^{d_2}$, Algorithm 1 computes the DMOC for a given (unstructured) grid for $[0, T_D]$. It is based on investigating all possible distances between points in $X_N$ and, therefore, has quadratic runtime.

**Algorithm 1** Evaluation of the DMOC on a general grid

**Input:** Data sites $X_N$, data values $Y_N$, grid $\mathcal{G}_T = \{t_1, \ldots, t_K\} \subset [0, T_D]$.
**Output:** $\omega_N(f_N, t_k)$, $k = 1, \ldots, K$

1: Set $\mathcal{I}_N := \{(i,j) : 1 \leq i < j \leq N\}$
2: **for** $k = 1, \ldots, K$ **do**
3:     Set $\omega_N(f_N, t_k) := 0$
4:     **for all** $(i,j) \in \mathcal{I}_N$ **do**
5:         Set $r := d_X(\boldsymbol{x}_i, \boldsymbol{x}_j),\ s := d_Y(\boldsymbol{y}_i, \boldsymbol{y}_j)$
6:         **for all** $k$ such that $r \leq t_k$ **do**
7:             Set $\omega_N(f_N, t_k) := \max\{\omega_N(f_N, t_k), s\}$
8:         **end for**
9:     **end for**
10: **end for**

We remark that the update in lines 6–8 can be performed in a lazy fashion with a constant time lookup such that the runtime of the algorithm is $\mathcal{O}(N^2 + K)$.

For large data sets, the quadratic runtime of Algorithm 1 is prohibitive and we resort to the minibatch version suggested in Algorithm 2, which assumes that the input data set is randomly shuffled. The

**Algorithm 2** Minibatch evaluation of the DMOC on a general grid

**Input:** Data sites $X_N$, data values $Y_N$, grid $\mathcal{G}_T = \{t_1, \ldots, t_K\} \subset [0, T_D]$, batch size $C$.
**Output:** $\widehat{\omega}_N(f_N, t_k)$, $k = 1, \ldots, K$

1: Set $\widehat{\omega}_N(f_N, t_k) := 0,\ k = 1, \ldots, K$
2: **for** $b = 1, \ldots, \lfloor \frac{N}{C} \rfloor$ **do**
3:     Set $\mathcal{I}_b := \{(i,j) : (b-1)C < i < j \leq bC\}$
4:     Set $\omega_b(f_N, t_k) := 0,\ k = 1, \ldots, K$
5:     **for all** $(i,j) \in \mathcal{I}_b$ **do**
6:         Set $r := d_X(\boldsymbol{x}_i, \boldsymbol{x}_j),\ s := d_Y(\boldsymbol{y}_i, \boldsymbol{y}_j)$
7:         **for all** $k$ such that $r \leq t_k$ **do**
8:             Set $\omega_b(f_N, t_k) := \max\{\omega_b(f_N, t_k), s\}$
9:         **end for**
10:         **for** $k = 1, \ldots, K$ **do**
11:             $\widehat{\omega}_N(f_N, t_k) := \max\{\widehat{\omega}_N(f_N, t_k), \omega_b(f_N, t_k)\}$
12:         **end for**
13:     **end for**
14: **end for**

algorithm has cost $\mathcal{O}(BC^2) = \mathcal{O}(NC)$, where $C$ is the batch size and $B = \lfloor N/C \rfloor$ the number of batches. The corresponding seminorm is easily computed by replacing $\omega_b$ by $\nu_\rho$, see Equation (1).

# 3 Experiments

## 3.1 Alignment metrics

To quantify the difference between different DMOCs, we introduce two alignment metrics. For two vectors $\boldsymbol{w}, \boldsymbol{w}' \in \mathbb{R}^K$, we define the relative alignment as $A(\boldsymbol{w}, \boldsymbol{w}') := \|\boldsymbol{w} - \boldsymbol{w}'\|_{\ell^1} / \|\boldsymbol{w}\|_{\ell^1}$. Clearly, as all norms on $\mathbb{R}^K$ are equivalent, different choices are possible. An alignment score can be computed from $A$ by considering

$$S(\boldsymbol{w}, \boldsymbol{w}') := 1 - A(\boldsymbol{w}, \boldsymbol{w}').$$

The score becomes negative if the relative alignment exceeds 1, which hints to very poor alignment. Moreover, we consider the Pearson correlation coefficient as a measure for alignment. Letting $\overline{\boldsymbol{w}} := (\|\boldsymbol{w}\|_{\ell^1}/K)\mathbf{1}$, we set

$$r(\boldsymbol{w}, \boldsymbol{w}') := \frac{(\boldsymbol{w} - \overline{\boldsymbol{w}})^\intercal (\boldsymbol{w}' - \overline{\boldsymbol{w}}')}{\|\boldsymbol{w} - \overline{\boldsymbol{w}}\|_{\ell^2} \|\boldsymbol{w}' - \overline{\boldsymbol{w}}'\|_{\ell^2}}.$$

We denote by $X^{\mathrm{u}} = X^{\mathrm{tr}} \cup X^{\mathrm{te}}$ the full data set comprising the training and test set. We further distinguish the site-to-value $f_X$ associated to $X$, from the untrained network $f_\theta$ and the trained network $f_{\widehat{\theta}}$. For the given function $f$ and the data sites $X$, we define:

$$S_{\text{untrained}} := S\big(\omega_{X^{\mathrm{u}}}(f_{X^{\mathrm{u}}}, t_k), \omega_{X^{\mathrm{u}}}(f_\theta, t_k)\big),\ r_{\text{untrained}} := r\big(\omega_{X^{\mathrm{u}}}(f_{X^{\mathrm{u}}}, t_k), \omega_{X^{\mathrm{u}}}(f_\theta, t_k)\big).$$

and, accordingly,

$$S_{\text{trained}} := S\big(\omega_{X^{\mathrm{u}}}(f_{X^{\mathrm{u}}}, t_k), \omega_{X^{\mathrm{u}}}(f_{\widehat{\theta}}, t_k)\big),\ r_{\text{trained}} := r\big(\omega_{X^{\mathrm{u}}}(f_{X^{\mathrm{u}}}, t_k), \omega_{X^{\mathrm{u}}}(f_{\widehat{\theta}}, t_k)\big)$$

## 3.2 Experimental setup

We compute the DMOC for different neural network architectures on the California housing data set [Pace and Barry, 1997], the Iris data set [Anderson, 1935, Fisher, 1936], the MNIST data set [LeCun, 1998] and the ImageNet data set [Deng et al., 2009]. As a baseline for comparison, we consider the DMOC of the site-to-value maps induced by the full data sets. Comparisons of the Lipschitz estimates from DMOC, against trivial upper bound and ECLipsE are provided. Following [Xu and Sivaranjani, 2024] the networks hereby considered will be feedforward networks consisting of affine maps and ReLU activation functions, hereafter termed as ReLU networks. An exception is made for the ImageNet data set, where the DMOC of AlexNet [Krizhevsky et al., 2012] is analyzed via the minibatch approach. The ReLU architectures considered include networks with 3, 5, and 20 layers, and widths of 50, 100, and 200 neurons. The networks $(3, 50)$, $(3, 100)$, and $(3, 200)$ are used to represent well-fitting regimes, while the deeper networks $(20, 50)$, $(20, 100)$, and $(20, 200)$ are designed to exhibit underfitting behavior. The architectures with 5 layers are constructed to reflect overfitting regimes. Within each class, similar DMOC pattern are observed. We therefore only report a representative example for each of the well-fitting, underfitting and overfitting cases. To obtain a good resolution near zero, an exponential grid with $K = 10^4$ is employed in both Algorithms 1 and 2. This choice also enables more accurate estimates of the Lipschitz constant derived from DMOC. In the subsequent plots, always a subsample of 100 grid points is shown, and logarithmic scaling is applied to both axes. Moreover, we exclusively consider the Euclidean metric here.

The baseline alignment score $S_{\text{data}}$, which compares the DMOC of the union data set to the DMOC on the training set, equals $9.92 \cdot 10^{-1}$, $1.00$ and $1.00$ for the California housing data set, the Iris data set and the MNIST data set, respectively, when comparing the DMOC of the training set to the DMOC of the union data set whereas the correlation coefficient $r_{\text{data}}$ takes values 0.99, 1.00 and 0.99. These results indicate that the training set already captures essentially all the smoothness properties required by the network to represent the union data set.

## 3.3 California housing

We first consider a one-layer neural network, consisting of an affine map, as in this setting the trivial bound from Remark 2.2 reduces to the exact Lipschitz constant of the network. The latter is given by the (Euclidean) operator norm of the weight matrix. The Lipschitz constant estimate provided by the DMOC is equal to $1.11$ compared to the exact constant $1.12$ provided by the operator norm.

Figure 1: DMOC for the California housing data set. We consider the DMOC of the union data set, for a trained and an untrained one-layer networks, across 100 weights initializations. The error bars correspond to one standard deviation. The dashed line indicates the Lipschitz constant computed using the DMOC.

Figure 1 shows the DMOC of the data, of the untrained network and the trained network together with the DMOC-Lipschitz constant of the trained network (dashed line). We have performed 100 random initialization of the network. The error bars correspond to one standard deviation. Across all experiments, the maximum absolute deviation of the Lipschitz constant estimate from the exact constant is $1.12 \cdot 10^{-2}$. At small scales, the DMOC of the untrained network shows more smoothness, while it underestimates the expected smoothness at larger scales. The DMOC of the trained network closely matches that of the data, though it tends to overestimate smoothness at small scales and underestimate it at larger ones.

The effect of nonlinear approximation becomes evident by a direct comparison of the DMOC of the one-layer network with the ReLU net $(3, 50)$, see left panel of Figure 2. Indeed the presence of nonlinearities enables the network to better capture the smoothness of data, exhibiting improved alignment with the DMOC of the data. Figure 2 further presents the DMOC of an overfitting network in the middle panel and of an underfitting network in the right panel. While the well-fitting network aligns with the DMOC on larger scales, the overfitting network overshoots while the underfitting network basically has no variation at any scale.

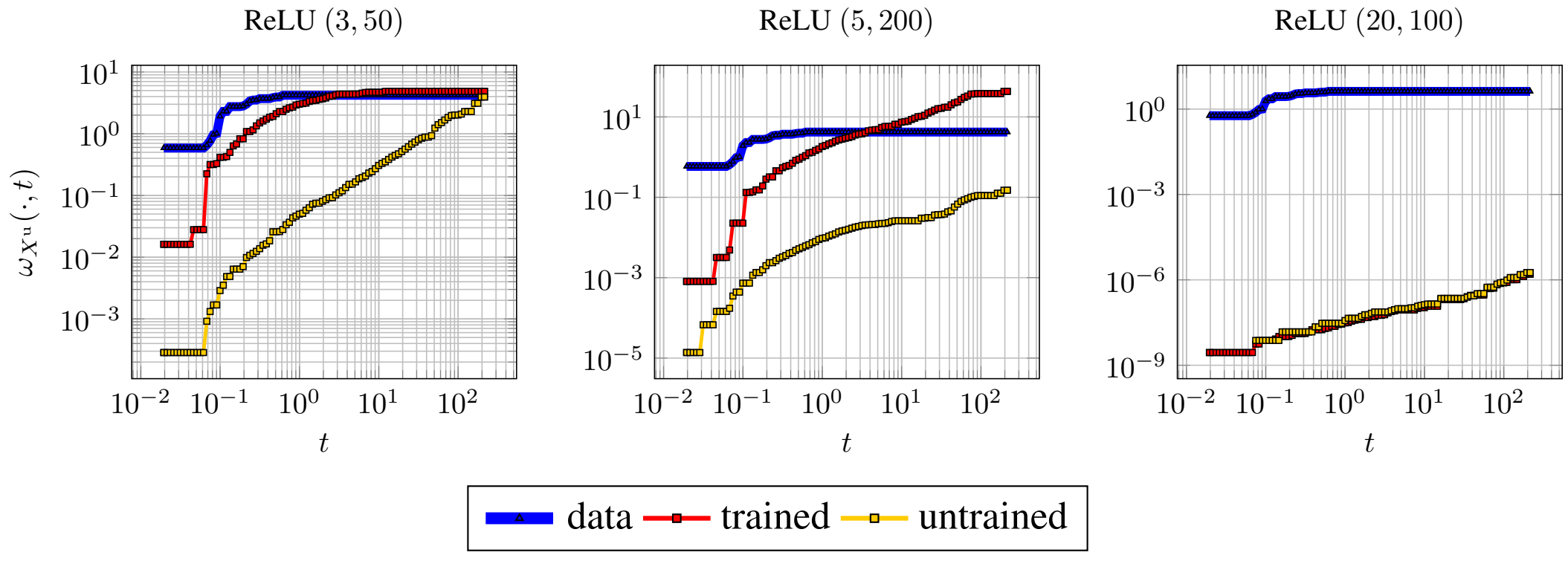

Figure 2: DMOC for the California housing data set. On the left, a well-fitting network is shown, while the middle panel shows overfitting and the right panel and underfitting network.

In Figure 3 we report the DMOC of well-fitting networks on the California housing data set, where the Lipschitz constant obtained from the DMOC is indicated by the slope of the dashed line. The solid line is the Lipschitz estimate by ECLipsE. In the presented cases, both estimates are nearly indistinguishable. However, for the networks $(3, 50), (3, 100), (3, 200), (5, 200)$, ECLipsE underestimates the Lipschitz constant. Indeed for all the considered networks ECLipsE is reported to rely on ECLipsE-Fast. A comprehensive comparison of the DMOC Lipschitz estimates against ECLipsE for the California housing data set is reported in Appendix C in Table 4. All our DMOC computations finished within 9 seconds, being faster than ECLipsE on five instances. Specifically for $(3, 200)$, $(3, 100)$, $(20, 200)$, $(5, 100)$, $(5, 200)$ where ECLipsE takes 30, 96, 17, 12 and 38 seconds respectively.

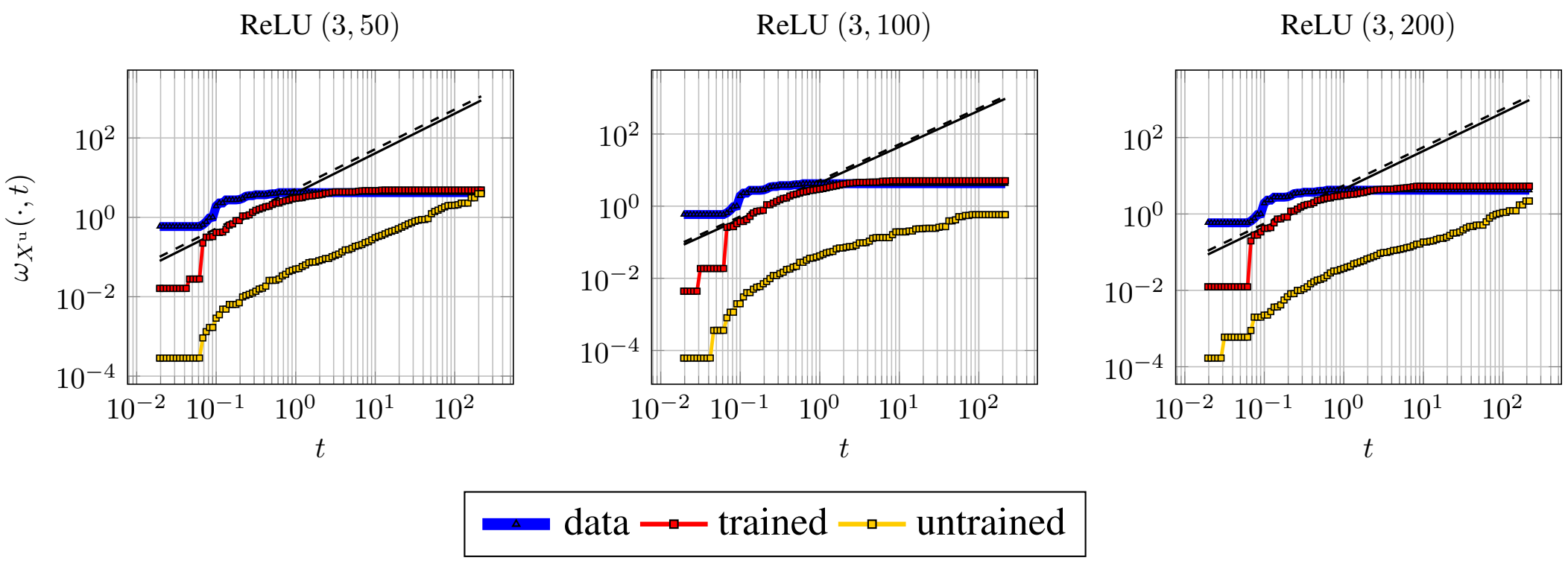

Figure 3: DMOC and Lipschitz constants computed using DMOC (dashed line) and ECLipsE (solid line) for the well-fitting network on the California housing data set.

To quantify the alignment between the DMOC of the data and the one of the neural networks, we employ the metrics from Subsection 3.1. Table 1, reports the alignment score $S$ and the Pearson correlation coefficient for the California housing data set. For the trained, well-fitting network, we observe a clearly better alignment with the DMOC of the data than for all the other cases.

Table 1: Alignment metrics on California data set for ReLU networks.

| | well-fitting | | | overfitting | | | underfitting | | |
|---|---|---|---|---|---|---|---|---|---|
| (layers, neurons/layer) | $(3, 50)$ | $(3, 100)$ | $(3, 200)$ | $(5, 50)$ | $(5, 100)$ | $(5, 200)$ | $(20, 50)$ | $(20, 100)$ | $(20, 200)$ |
| $S_{\text{trained}}$ | $7.35 \cdot 10^{-1}$ | $7.06 \cdot 10^{-1}$ | $6.88 \cdot 10^{-1}$ | $5.35 \cdot 10^{-8}$ | $5.62 \cdot 10^{-8}$ | $1.28 \cdot 10^{-8}$ | $-1.15$ | $-5.75 \cdot 10^{-1}$ | $-1.29$ |
| $S_{\text{untrained}}$ | $1.47 \cdot 10^{-1}$ | $4.76 \cdot 10^{-2}$ | $7.81 \cdot 10^{-2}$ | $4.98 \cdot 10^{-8}$ | $6.40 \cdot 10^{-8}$ | $1.71 \cdot 10^{-8}$ | $1.53 \cdot 10^{-2}$ | $1.28 \cdot 10^{-2}$ | $8.44 \cdot 10^{-3}$ |
| $r_{\text{trained}}$ | $8.62 \cdot 10^{-1}$ | $8.57 \cdot 10^{-1}$ | $8.55 \cdot 10^{-1}$ | $4.50 \cdot 10^{-1}$ | $3.35 \cdot 10^{-1}$ | $5.36 \cdot 10^{-1}$ | $4.46 \cdot 10^{-1}$ | $4.78 \cdot 10^{-1}$ | $4.28 \cdot 10^{-1}$ |
| $r_{\text{trained}}$ | $3.52 \cdot 10^{-1}$ | $4.89 \cdot 10^{-1}$ | $3.53 \cdot 10^{-1}$ | $4.53 \cdot 10^{-1}$ | $3.33 \cdot 10^{-1}$ | $4.56 \cdot 10^{-1}$ | $3.40 \cdot 10^{-1}$ | $3.66 \cdot 10^{-1}$ | $4.50 \cdot 10^{-1}$ |

### 3.4 Classifiers on Iris and MNIST

In classification settings with one-hot encoded labels and Euclidean distance, the DMOC of the site-to-value map is a piecewise constant function with a single jump, which in turn provides an insightful interpretation of the network's DMOC. Specifically, the DMOC is equal to zero up to a certain scale $t'$ and equal to $\sqrt{2}$ afterwards. The value $\sqrt{2}$ corresponds to the maximum Euclidean distance between different one-hot labels. The distance $t'$ is the minimum cross-class distance in the data set, i.e., the smallest distance between two points belonging to different classes. This yields a clear interpretation of the neural network's DMOC, as any function that tends to be consistent with the data DMOC must satisfy two competing constraints. For any grid site $t < t'$, the DMOC should remain small, and very close to zero, since same-class points should map to the same label. For $t \geq t'$, the DMOC should reach $\sqrt{2}$, because there exist pairs at distance at most $t'$ with different labels that must be mapped sufficiently far apart.

Figure 4 reports the DMOC of well-fitting, overfitting and underfitting ReLU networks on the Iris data set both before and after training, revealing a consistent pattern. The DMOC of the well trained network assumes lower values before the minimum cross-class distance, then aligning with the $\sqrt{2}$ bound, coherently with the aforementioned pattern. Secondly, well trained networks, once optimized, align closely with the reference DMOC of the data. In contrast, underfitting networks exhibit substantial misalignment. Overfitting models may approach or partially align with the DMOC of the data, but frequently overshoot or undershoot smoothness across multiple regions.

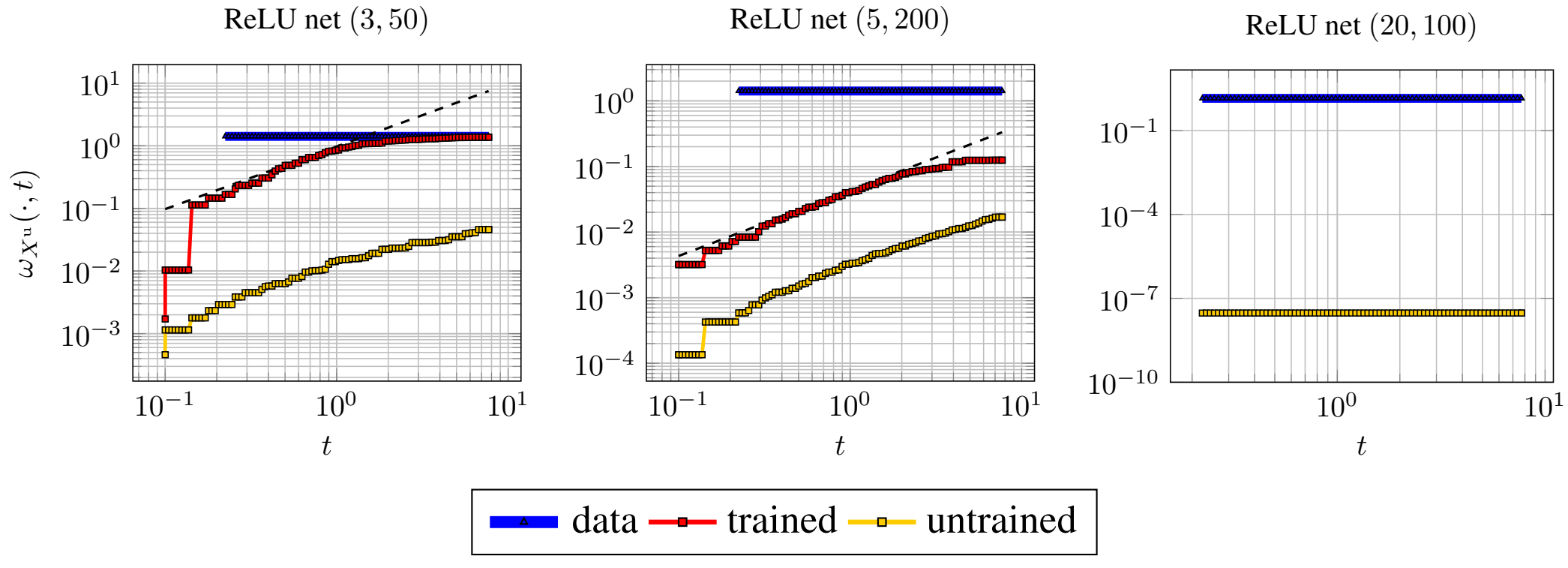

Figure 4: DMOC for the Iris data set. On the left, a well-fitting network is shown, while the middle panel shows overfitting and the right panel refers to an underfitting network. The Lipschitz estimate from DMOC for the networks corresponds to the slope of the respective dashed lines.

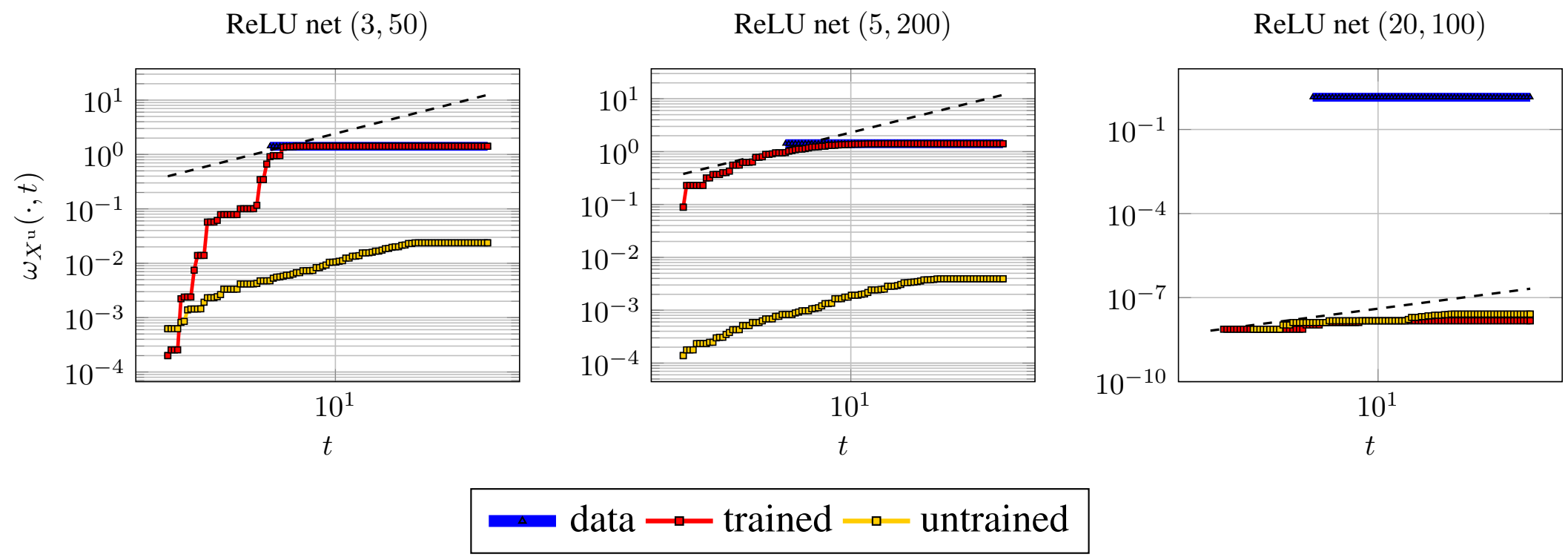

Figure 5: DMOC for MNIST data set. On the left, a well-fitting network is shown, while the middle panel shows overfitting and the right panel refers to an underfitting network. The Lipschitz estimate from DMOC for the networks corresponds to the slope of the respective dashed lines.

Table 2 reports the alignment metrics for the considered ReLu networks on the Iris data set. As before, the DMOC of well-fitting networks is strongly correlated to the DMOC of the data. This trend is corroborated by results on the MNIST dataset, see 3, suggesting that the phenomenon is not limited to current settings. Specifically, the alignment metrics reported in Table 3 suggest stronger linear correlation between DMOCs of site-to-value map and well-fitting networks.

Table 2: Alignment metrics on Iris data set for ReLU networks. The * denotes a networks with a constant output, and a constant DMOC.

| | well-fitting | | | overfitting | | | underfitting | | |
|---|---|---|---|---|---|---|---|---|---|
| (layers, neurons/layer) | $(3, 50)$ | $(3, 100)$ | $(3, 200)$ | $(5, 50)$ | $(5, 100)$ | $(5, 200)$ | $(20, 50)$ | $(20, 100)$ | $(20, 200)$ |
| $S_{\text{trained}}$ | $6.16 \cdot 10^{-1}$ | $7.19 \cdot 10^{-1}$ | $7.98 \cdot 10^{-1}$ | 0.0 | 0.0 | $1.83 \cdot 10^{-8}$ | $2.97 \cdot 10^{-3}$ | $3.28 \cdot 10^{-2}$ | $4.18 \cdot 10^{-2}$ |
| $S_{\text{untrained}}$ | $1.29 \cdot 10^{-2}$ | $2.08 \cdot 10^{-2}$ | $8.07 \cdot 10^{-3}$ | 0.0 | $2.11 \cdot 10^{-8}$ | $2.47 \cdot 10^{-8}$ | $3.13 \cdot 10^{-3}$ | $1.91 \cdot 10^{-3}$ | $4.01 \cdot 10^{-3}$ |
| $r_{\text{trained}}$ | $6.43 \cdot 10^{-1}$ | $7.21 \cdot 10^{-1}$ | $7.84 \cdot 10^{-1}$ | * | * | $6.17 \cdot 10^{-1}$ | $4.92 \cdot 10^{-1}$ | $5.21 \cdot 10^{-1}$ | $5.05 \cdot 10^{-1}$ |
| $r_{\text{untrained}}$ | $5.08 \cdot 10^{-1}$ | $4.54 \cdot 10^{-1}$ | $5.41 \cdot 10^{-1}$ | * | 1.0 | $6.86 \cdot 10^{-1}$ | $4.16 \cdot 10^{-1}$ | $5.87 \cdot 10^{-1}$ | $4.41 \cdot 10^{-1}$ |

### 3.5 Minibatch approach for ImageNet

For ImageNet, we compute the DMOC on the union of the training set and the validation set using the minibatch algorithm. Figure 6 shows the resulting DMOC for the untrained AlexNet (left panel) and the trained AlexNet (right panel) for increasing batch sizes. The DMOCs of the trained network are significantly larger than those of the untrained network, while an increasing resolution of the DMOC closer to zero is observed for increasing batch sizes. The Lipschitz constant varies with the batch size and corresponds to $7.70 \cdot 10^{-3}$, $1.06 \cdot 10^{-2}$, $1.06 \cdot 10^{-2}$ respectively for $C = 10, 100, 1\,000$. Computation time spans from 120 minutes for $C = 10$ up to 442 minutes for $C = 1\,000$.

Table 3: Alignment metrics on MNIST data set for ReLU networks.

| | well-fitting | | | overfitting | | | underfitting | | |
|---|---|---|---|---|---|---|---|---|---|
| (layers, neurons/layer) | $(3, 50)$ | $(3, 100)$ | $(3, 200)$ | $(5, 50)$ | $(5, 100)$ | $(5, 200)$ | $(20, 50)$ | $(20, 100)$ | $(20, 200)$ |
| $S_{\text{trained}}$ | $9.50 \cdot 10^{-1}$ | $9.49 \cdot 10^{-1}$ | $9.45 \cdot 10^{-1}$ | $1.83 \cdot 10^{-8}$ | $7.97 \cdot 10^{-9}$ | $3.44 \cdot 10^{-9}$ | $5.38 \cdot 10^{-1}$ | $8.62 \cdot 10^{-1}$ | $7.82 \cdot 10^{-1}$ |
| $S_{\text{untrained}}$ | $1.05 \cdot 10^{-2}$ | $8.76 \cdot 10^{-3}$ | $1.10 \cdot 10^{-2}$ | $8.40 \cdot 10^{-9}$ | $1.24 \cdot 10^{-8}$ | $1.93 \cdot 10^{-8}$ | $1.79 \cdot 10^{-3}$ | $1.54 \cdot 10^{-3}$ | $1.76 \cdot 10^{-3}$ |
| $r_{\text{trained}}$ | $9.82 \cdot 10^{-1}$ | $9.70 \cdot 10^{-1}$ | $9.72 \cdot 10^{-1}$ | $6.10 \cdot 10^{-1}$ | $9.01 \cdot 10^{-1}$ | $4.13 \cdot 10^{-1}$ | $8.18 \cdot 10^{-1}$ | $9.17 \cdot 10^{-1}$ | $9.09 \cdot 10^{-1}$ |
| $r_{\text{untrained}}$ | $7.25 \cdot 10^{-1}$ | $7.24 \cdot 10^{-1}$ | $7.29 \cdot 10^{-1}$ | $7.39 \cdot 10^{-1}$ | $7.87 \cdot 10^{-1}$ | $4.50 \cdot 10^{-1}$ | $7.45 \cdot 10^{-1}$ | $7.56 \cdot 10^{-1}$ | $7.31 \cdot 10^{-1}$ |

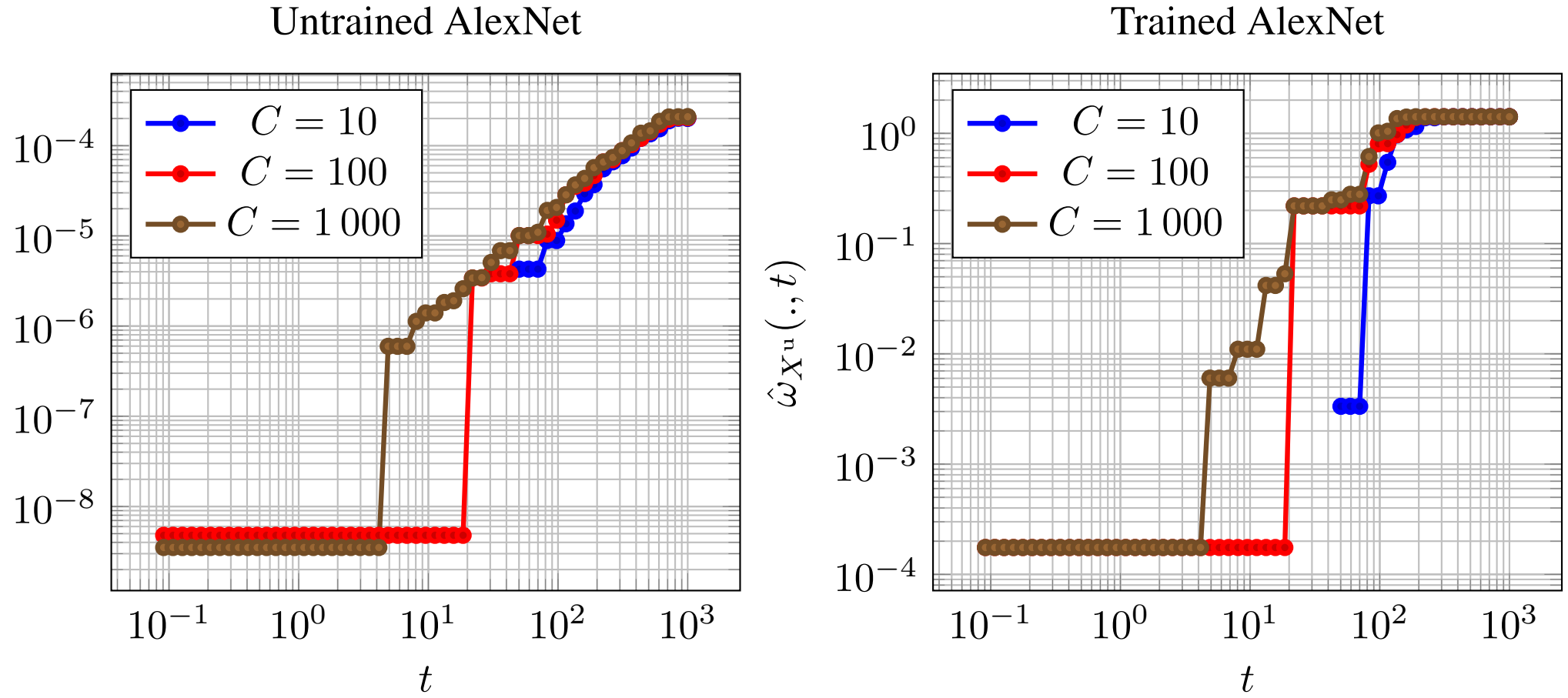

Figure 6: Minibatch DMOC for AlexNet trained on the ImageNet data set. The left plot reports the DMOC for the untrained network and increasing batch sizes, while the DMOC of the trained network is shown for different batch sizes on the right.

# 4 Conclusion

The DMOC shifts the focus from model-centric notions of regularity to regularity relative to the data. This yields a multi-scale black-box characterization of input-output behavior and provides a more refined description than the Lipschitz constant. The DMOC is computed from input-output pairs in metric spaces and is therefore architecture agnostic. However, its reliability is intrinsically tied to the geometric quality of the underlying data, as reflected by Theorem 2.4.

Theorem 2.4 establishes finite-sample convergence of DMOC-induced seminorms to their continuous counterparts. The explicit non-asymptotic bounds depend only on the modulus of continuity of the target function and the underlying data set, quantified by separation and fill distance. We have proposed an exact algorithm with quadratic cost for computing the DMOC and a more efficient minibatch version that reduces the cost to evaluations within batch pairs. Empirically, we have benchmarked the approach on feedforward and convolutional architectures across the California Housing, Iris, MNIST, and ImageNet data sets. We have demonstrated empirically that the DMOC profiles offer a richer multi-scale description of network regularity than the Lipschitz constant. In addition, the proposed alignment metrics allow automated comparison of DMOCs. The obtained DMOC profiles clearly distinguish trained from untrained networks and reveal characteristic signatures of underfitting and overfitting, corresponding respectively to overly smooth behavior and oscillatory deviations relative to the data. As a byproduct, we obtain Lipschitz estimates that are competitive with ECLipsE and ECLipsE-Fast.

We conclude by noting two limitations of the present work. First, although the general metric-space framework makes DMOC directly applicable to latent representations used in transformer networks, this has not yet been explored empirically. Extending the experimental evaluation to contemporary architectures is a natural and important next step. Second, the current work uses DMOC exclusively as a post hoc diagnostic. An important direction for future work is to incorporate DMOC into training, for example through regularization terms that explicitly shape the smoothness profile of a network.

# Appendices

## A Technical Proofs

**Proof of Theorem 2.4**.

*Proof.* We define $|f|_{\rho,N} := \max_{t\in Z_N\setminus\{0\}} \nu_\rho(f,t)$. Then, we split

$$0 \le |f|_\rho - |f_N|_{\rho,N} \le \underbrace{|f|_\rho - |f|_{\rho,N}}_{=:(\spadesuit)\ge 0} + \underbrace{|f|_{\rho,N} - |f_N|_{\rho,N}}_{=:(\clubsuit)\ge 0}$$

and estimate both differences separately. To estimate ($\spadesuit$), we observe that $Z_N$ is an $r$-net for $[0,T_D]$, see [Dölz and Multerer, 2025, Lemma 3.4]. Thus, there is a partition $\{P_t\}_{t\in Z_N}$ of $[0,T_D]$ into mutually disjoints sets with $t\in P_t$ and $\operatorname{diam}(P_t) < 2r$ for all $t\in Z_N$. Setting $\Xi_{Z_N} = \{(t,P_t)\}_{t\in Z_N}$ allows to introduce the interpolation operator

$$I_{\Xi_{Z_N}} g := \sum_{t\in Z_N} g(t)\mathbb{1}_{P_t}$$

acting on continuous functions $g\colon [0,T_D]\to\mathbb{R}$. Herein, $\mathbb{1}_{P_t}$ denotes the indicator function of the set $P_t$. The approximation result [Dölz and Multerer, 2025, Theorem 6.1] yields

$$(\spadesuit) = |f|_\rho - |f|_{\rho,N} = \|\nu_\rho(f,\cdot) - I_{\Xi_{Z_N}}\nu_\rho(f,\cdot)\|_{C([0,T_D])} \le \omega\big(\nu_\rho(f,\cdot),2r\big).$$

To estimate ($\clubsuit$), consider

$$\begin{aligned}(\clubsuit) = |f|_{\rho,N} - |f_N|_{\rho,N} &= \max_{z\in Z_N\setminus\{0\}} \nu_\rho(f,z) - \max_{z\in Z_N\setminus\{0\}} \nu_{\rho,N}(f_N,z)\\ &\le \max_{z\in Z_N\setminus\{0\}} \rho(z)^{-1}\big(\omega(f,z) - \omega_N(f_N,z)\big).\end{aligned}$$

Now, observe that

$$0 \le \omega(f,z) - \omega_N(f_N,z) \le \begin{cases}\omega(f,z) & z\le r,\\ 3\omega(f,r) & z > r,\end{cases}$$

due to $\omega_N(f_N,z) = 0$ for $z\le r$ and Theorem 2.3 for $z > r$. The approximation result [Dölz and Multerer, 2025, Corollary 3.10] with the choice $\alpha = r$ yields

$$\begin{aligned}(\clubsuit) &\le \max_{z\in Z_N\setminus\{0\}} \rho(z)^{-1}\big(\omega(f,z) - \omega_N(f_N,z)\big) \le \rho(q_N)^{-1}\|\omega(f,\cdot) - \omega_N(f_N,\cdot)\|_{C([0,T_D])}\\ &\le \rho(q_N)^{-1}\big(3\omega(f,r) + \|\omega(f,\cdot) - \omega(f,\cdot - r)\|_{C([r,T_D])}\big).\end{aligned}$$

Exploiting subadditivity of the MOC yields $0 \le \omega(f,t) - \omega(f,t-r) \le \omega(f,r)$, for all $t\ge r$ and thus the assertion. $\square$

## B Training setup

All experiments in this paper have been carried out on a MacBook Pro M4 Max with 36GB of unified memory. For all networks the number of neurons is fixed to be the same across all layers.

**California:** All linear models have been trained with Mean Square Error (MSE) loss on a 70-20 train-test split, with 100 epochs achieving an $R^2$ of at least 0.6 on the training set and of 0.58 on test set. The ReLU networks $(3, 50), (3, 100), (3, 200)$ achieve an $R^2$ of at least 0.70 similar to the linear case. While the three networks $(20, 50), (20, 100), (20, 200)$ all achieve an $R^2$ near zero on both training set and test set. The three remaining models consisting of five layers, all achieve an $R^2$ of at least 0.92 on the training set, while a near zero $R^2$ on the test set.

**Iris:** The three models $(3, 50), (3, 100), (3, 200)$ well-trained achieve an accuracy of at least 0.96 on the training set, and and accuracy of 1 on the test set. Networks $(20, 50), (20, 100), (20, 200)$ achieve an accuracy of 0.33 on the training set and on the test set. The three remaining five layer models achieve an accuracy of at least 0.80 on the training set and an accuracy of 0.33 on the test set. All the models were trained on 70-20 train-test split.

**MNIST:** The three models $(3, 50), (3, 100), (3, 200)$ are well-trained and achieve an accuracy of at least 0.96 on the training set and of 1 on the test set. Networks $(20, 50), (20, 100), (20, 200)$ achieve an accuracy of 0.33 on the training set and on the test set. The three remaining five layer models achieve an accuracy of at least 0.80 on the training set and of 0.33 on the test set. The networks were trained on 60.000 training images, performances have been evaluated on 10 000 test images.

**ImageNet:** For ImageNet the data set consists of a union of 50 000 data points from the validation set and of 1 500 000 data points from the training set. The union is hereby given by a stack of all chunks of size $C$ from training set and test set, excluding any leftover samples so that all batches are evenly sized. A pretrained AlexNet with weights from TorchVision is used. The network achieves approximately 0.56 top-1 accuracy and 0.79 top-5 accuracy on the ImageNet validation set.

## C Additional Experimental Results

**Lipschitz estimates** Table 4 shows the computed bounds on the Lipschitz constant with the trivial bound from Remark 2.2 as well as ECLipsE and DMOC for trained ReLU networks on California housing data set.

Table 4: Lipschitz constant comparison for ReLU networks on the California housing data set.

| | well-fitting | | | overfitting | | | underfitting | | |
|---|---|---|---|---|---|---|---|---|---|
| (layers, neurons/layer) | $(3,50)$ | $(3,100)$ | $(3,200)$ | $(5,50)$ | $(5,100)$ | $(5,200)$ | $(20,50)$ | $(20,100)$ | $(20,200)$ |
| Trivial | $1.34 \cdot 10^{2}$ | $3.72 \cdot 10^{2}$ | $1.01 \cdot 10^{3}$ | $1.53 \cdot 10^{3}$ | $9.09 \cdot 10^{3}$ | $3.88 \cdot 10^{4}$ | $1.04 \cdot 10^{12}$ | $1.06 \cdot 10^{15}$ | $1.00 \cdot 10^{18}$ |
| ECLipsE | 4.10 | 4.41 | 4.45 | 2.05 | 1.99 | 1.80 | $5.69 \cdot 10^{-2}$ | $5.46 \cdot 10^{-2}$ | $5.83 \cdot 10^{-2}$ |
| $L_{\text{DMOC}}$ | 5.22 | 5.17 | 5.60 | 1.81 | 1.75 | 1.95 | $1.26 \cdot 10^{-7}$ | $1.43 \cdot 10^{-7}$ | $6.61 \cdot 10^{-8}$ |

**Minibatch DMOC** We have tested the minibatch approach on the full MNIST data set for the well-trained ReLU network $(3,50)$. Figure 7 shows the convergence of the DMOC for increasing batch sizes towards the DMOC evaluated at all data sites.

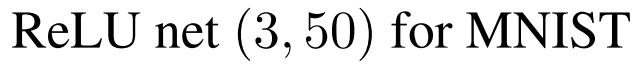

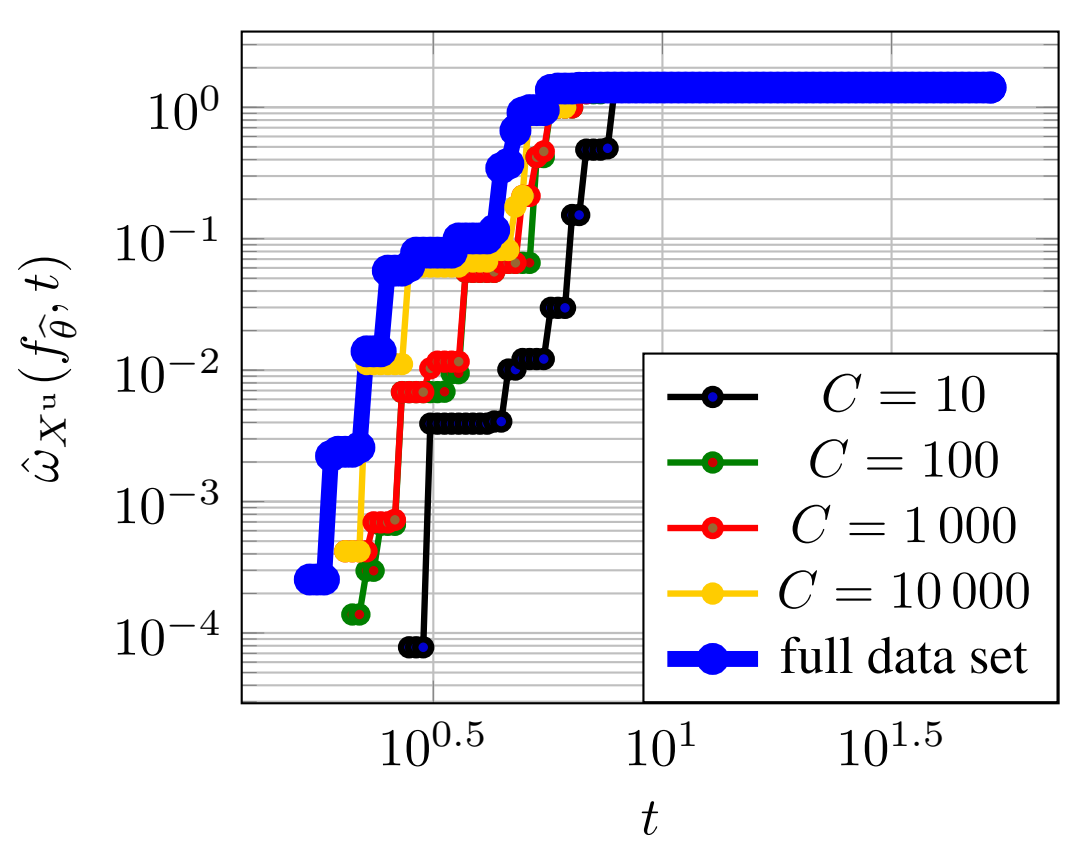

Figure 7: Minibatch DMOC convergence on the MNIST data set for ReLU net $(3,50)$ with respect to different batch sizes $C$

We further report the The Lipschitz constants for trained network obtained from DMOC across varying batch sizes according to $1.79 \cdot 10^{-1}$, $1.97 \cdot 10^{-1}$, $1.97 \cdot 10^{-1}$, $2.15 \cdot 10^{-1}$, $2.42 \cdot 10^{-1}$, respectively, for batch size $C = 10, 100, 1\,, 10\,00$ and the full data set ($C = 70\,000$).